\pdfoutput=1

\documentclass[11pt]{article}

\usepackage[final]{acl}

\usepackage{times}
\usepackage{latexsym}
\usepackage{soul}
\usepackage[T1]{fontenc}

\usepackage[utf8]{inputenc}

\usepackage{microtype}

\usepackage{inconsolata}

\usepackage{graphicx}

\usepackage[utf8]{inputenc} 
\usepackage[T1]{fontenc}    
\usepackage{hyperref}       
\usepackage{url}            
\usepackage{booktabs}       
\usepackage{amsfonts}       
\usepackage{nicefrac}       
\usepackage{microtype}      
\usepackage{xcolor}         
\usepackage{graphicx} 
\usepackage{tikz}
\usepackage{graphicx}
\usepackage{algorithm}
\usepackage{algpseudocode}
\usepackage{caption}
\usepackage{amsmath} 
\usepackage{wrapfig}

\usepackage[T1]{fontenc}
\usepackage{hyperref}
\usepackage{url}
\usepackage{booktabs}
\usepackage{amsfonts}
\usepackage{nicefrac}
\usepackage{enumerate}
\usepackage{dialogue}
\usepackage{tcolorbox}
\tcbuselibrary{breakable}
\usepackage{listings}
\usepackage{ulem}
\usepackage{marvosym}
\usepackage{amssymb}
\usepackage{natbib}
\usepackage{url}            
\usepackage{booktabs}       
\usepackage{amsfonts}       
\usepackage{nicefrac}       

\title{\texttt{DreamFactory}: Pioneering Multi-Scene Long Video Generation with a Multi-Agent Framework}
\author{
  Zhifei Xie$^\spadesuit$ \textsuperscript{\Letter} \qquad \qquad
  Daniel Tang$^\clubsuit$ \qquad \qquad
  Dingwei Tan$^\diamondsuit$   \\
  \textbf{Jacques Klein}$^\clubsuit$ \qquad \qquad
  \textbf{Tegawendé F. Bissyandé}$^{\clubsuit}$ \qquad \qquad  \textbf{Saad Ezzini}$^{\heartsuit\clubsuit}$ \\
  $^\spadesuit$TsingHua University \qquad
  $^\clubsuit$University of Luxembourg \qquad
  $^\diamondsuit$Beijing Institute of Technology  
  \texttt{filicos.thu@gmail.com} \qquad 
  \texttt{xunzhu.tang@uni.lu}\qquad \qquad 
  \texttt{tdw03edu@outlook.com}\\
  \texttt{jacques.klein@uni.lu}\qquad
  \texttt{tegawende.bissyande@uni.lu}\qquad
  \texttt{s.ezzini@lancaster.ac.uk}
}

\begin{document}
\maketitle

\begin{abstract}
Current video generation models excel at creating short, realistic clips, but struggle with longer, multi-scene videos. We introduce \texttt{DreamFactory}, an LLM-based framework that tackles this challenge. \texttt{DreamFactory} leverages multi-agent collaboration principles and a Key Frames Iteration Design Method to ensure consistency and style across long videos. It utilizes Chain of Thought (COT) to address uncertainties inherent in large language models. \texttt{DreamFactory} generates long, stylistically coherent, and complex videos. Evaluating these long-form videos presents a challenge. We propose novel metrics such as Cross-Scene Face Distance Score and Cross-Scene Style Consistency Score. To further research in this area, we contribute the Multi-Scene Videos Dataset containing over 150 human-rated videos. \texttt{DreamFactory}\footnote{We will make our framework and datasets public after paper acceptance.} paves the way for utilizing multi-agent systems in video generation.

\end{abstract}

\section{Introduction}
Video, integrating both visual and auditory modalities—the most direct sensory pathways through which humans perceive and comprehend the world—effectively conveys information with compelling persuasiveness and influence, progressively becoming a powerful tool and medium for communication [\cite{tang1992users},~\cite{owen1992video},~\cite{armes2006video},~\cite{harris2016video},~\cite{merkt2011learning}]. Traditional video production is an arduous and time-intensive process, particularly for capturing elusive real-life scenes. Owing to the rapid advancements in deep learning, AI-driven video generation techniques now facilitate the acquisition of high-quality images and video segments with ease [~\cite{PIKA},~\cite{blattmann2023stable},~\cite{SORA},~\cite{blattmann2023align},~\cite{Runway},~\cite{gu2023reuse}]. However, crafting practical, multi-scene videos that meet real-world needs remains a formidable challenge. This includes ensuring consistency in character portrayal, stylistic coherence, and background across different scenes, proficiently maneuvering professional linguistic tools, and managing complex production steps beyond merely assembling brief video clips generated by current technologies. Therefore, there is an urgent need within the field of video generation for a model capable of directly producing long-duration, high-quality videos with high consistency, thus enabling AI-generated video to gain widespread acceptance and become a premier producer of content for human culture and entertainment. \looseness=-1

At the current stage, substantial advancements in the video domain utilize diffusion-based generative models, achieving excellent visual outcomes [~\cite{blattmann2023stable},~\cite{Runway},~\cite{SORA}]. Nonetheless, due to the intrinsic characteristics of diffusion models, the videos produced are typically short segments, usually limited to four seconds. For generating longer videos, models like LSTM and GANs are employed \cite{gupta2022rv}, however, these models struggle to meet the demands for high image quality and are restricted to synthesizing videos of lower resolution. These state-of-the-art approaches attempt to use a single model to address all sub-challenges of video generation end-to-end, encompassing attractive scriptwriting, character definition, and artistic shot design. However, these tasks are typically collaborative and not the sole responsibility of a single model. 

In addressing complex tasks and challenges in problem-solving and coding, researchers have begun utilizing LLM multi-agent collaborative techniques, modeled on human cooperative behaviors, and have observed numerous potent agents. With the integration of large models that include visual capabilities, multi-agent collaborative technologies have now developed an AI workflow capable of tackling challenges in the image and video domain. 

\noindent 

\begin{figure}
    \centering
    \includegraphics[width=1\linewidth]{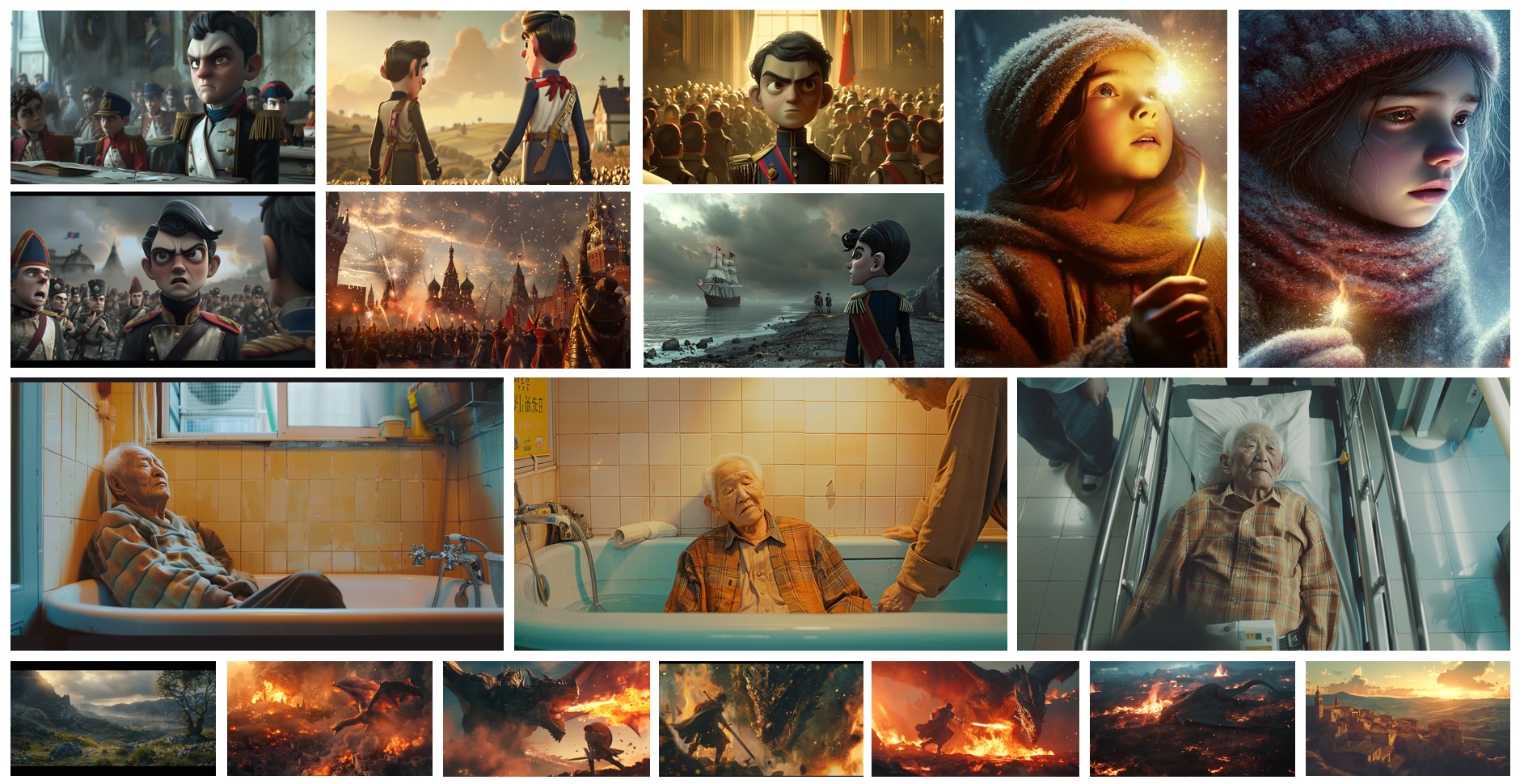}
    \caption{Keyframe data produced by \textbf{\texttt{DreamFactory}}. It can be seen that the character's facial features, visual style, and even clothing are consistent. }
    \label{fig:enter-label}
\end{figure}

In this paper, we introduce multi-agent collaborative techniques to the domain of video generation, developing a multi-scene long video generation framework named \textbf{\texttt{DreamFactory}}, which simulates an AI virtual film production team. Agents based on LLMs assume roles akin to directors, art directors, screenwriters, and artists, collaboratively engaging in scriptwriting, storyboard creation, character design, keyframe development, and video synthesis. We define the concept of keyframe in the long video generation field to maintain consistency across video segments. In \textbf{\texttt{DreamFactory}}, we draw on the successful CoT concept from the multi-agent reasoning process to devise a keyframe iteration method specific to video. To address the drift phenomenon in large language models, a Monitor role is introduced to ensure consistency between different frames. \textbf{\texttt{DreamFactory}} also establishes an integrated image vector database to maintain the stability of the creative process. Based on the algorithms discussed, \texttt{DreamFactory} can automate the production of multi-scene videos of unrestricted length with consistent image continuity.

To evaluate our framework, we employed state-of-the-art video generation models as tools, measuring video generation 
performance on the UTF-101 and HMDB51 datasets. Furthermore, given the novelty of our task, with few prior ventures 
into this area, we compared long videos generated by our framework against those produced using the original tools. 
We found that our model significantly outperformed the existing native models regarding evaluation mechanisms. 
Finally, we collected AI-generated short videos currently available on the internet and assessed them using mechanisms 
such as the Inception Score, alongside evaluations conducted by human judges. Our findings indicate that our videos 
surpass the average quality of those produced manually. Some examples generated by the framework are shown in Figure~\ref{fig:enter-label}.


\section{Related work}

\textbf{LLM-based Agents.} In recent years, the capabilities of large language models have been continually enhanced, exemplified by advancements such as GPT-4~\cite{GPT-4}, Claude-3~\cite{Claude-3}, and LLama-2~\cite{LLama-2}, among others. Subsequently, exploration into enhancing the abilities of these large language models has emerged, introducing methodologies such as CoT~\cite{wei2022chain}, ToT~\cite{yao2024tree}, ReACT~\cite{yao2022react}, Reflexion~\cite{shinn2024reflexion}, and various other approaches to facilitate iterative output and correction cycles. Within this context, the notion of Multi-agents has surfaced, with early research efforts including notable works such as Camel~\cite{li2024camel}, Voyager~\cite{wang2023voyager}, MetaGPT~\cite{hong2023metagpt}, ChatDev~\cite{qian2023communicative}, and AutoGPT~\cite{yang2023auto}. Recently, powerful Multi-agents frameworks have proliferated across diverse domains, with prominent instances in fields such as coding, including notable contributions such as CodeAgent~\cite{tang2024collaborative}, CodeAct~\cite{wang2024executable}, and Codepori~\cite{rasheed2024codepori}. Utilitarian tools such as Toolformer~\cite{schick2024toolformer}, HuggingGPT~\cite{shen2024hugginggpt}, Toolllm~\cite{qin2023toolllm}, and WebGPT~\cite{nakano2021webgpt} have also been employed. Other noteworthy endeavors encompass projects like WebArena~\cite{zhou2023webarena}, RET-LLM~\cite{modarressi2023ret}, and OpenAGI~\cite{ge2024openagi}, each contributing to the advancement and proliferation of Multi-agents paradigms.

\textbf{Video synthesis. }In the field of video generation, traditional methods primarily utilize Generative Adversarial Networks (GANs) for video creation, as demonstrated in the works of Tim Brooks et al.~\cite{brooks2022generating} and the foundational contributions of Ian Goodfellow et al.~\cite{goodfellow2014generative} However, in recent years, a significant shift has occurred towards leveraging the potent capabilities of diffusion processes, with pioneering research conducted by Jascha et al.~\cite{esser2023structure}, and Song et al.~\cite{song2020score}. The forefront of this evolution is marked by the development of Latent Video Diffusion Models. This approach is exemplified in the seminal efforts of Andreas Blattmann et al.~\cite{blattmann2023align}, Gu et al.~\cite{gu2023reuse},Guo et al.~\cite{guo2023animatediff}, He et al.~\cite{he2022latent} and Wang et al.~\cite{wang2023modelscope}. Currently, the most formidable advancements in this area are four main models: Pika~\cite{PIKA}, Stable Video~\cite{blattmann2023stable}, Runway~\cite{Runway}, and Sora~\cite{SORA}.

\begin{figure*}[t]
  \centering
  \includegraphics[width=1\textwidth]{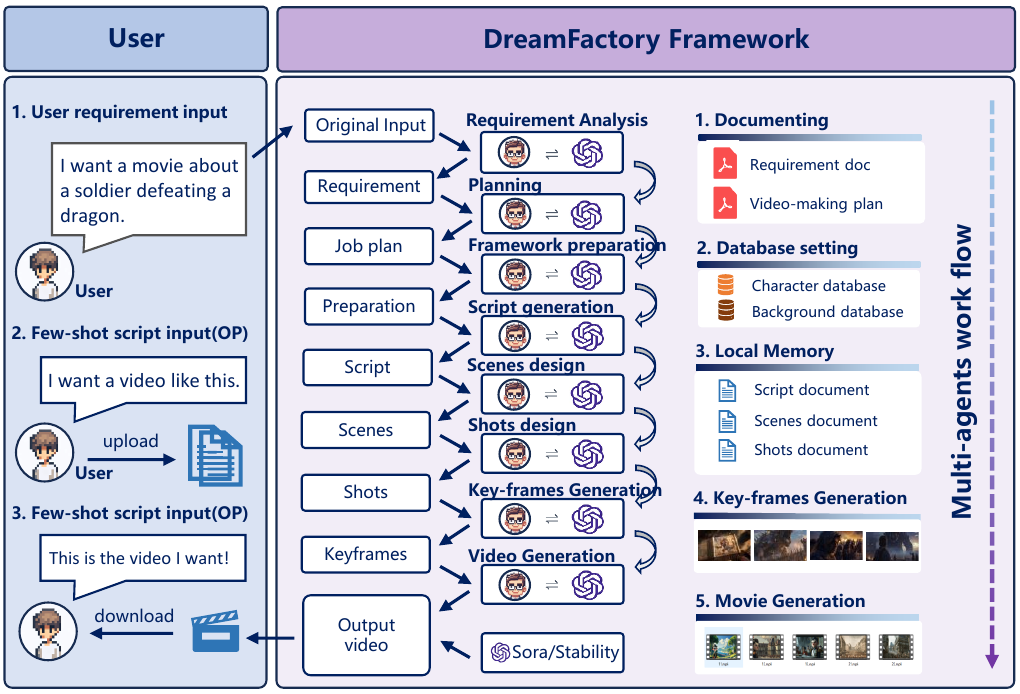}
  \caption{An overview of the \texttt{DreamFactory} framework. The framework transposes the entire filmmaking process into AI, forming an AI-driven video production team.}
  \label{fig:overview}
\end{figure*}
\section{DreamFactory}
Our \texttt{DreamFactory} framework utilizes multiple large language models (LLMs) to form a simulated animation company, 
taking on roles such as CEO, Director, and Creator. Given a story, they collaborate and create a video through social 
interaction and cooperation. This framework allows LLMs to simulate the real world by using small video generation 
models as tools to accomplish a massive task. 
This section details the methodology behind our innovative \texttt{DreamFactory} framework. We first describe the defined role 
cards in Section 3.1 and discuss the pipeline in Section 3.2. Finally, we will discuss the keyframe iteration design method.

\subsection{Role Definition}

In the architecture of our simulation animation company \textbf{\texttt{DreamFactory}}, the following roles are included: CEO, movie director, 
film producer, Screenwriter, Filmmaker, and Reviewer.Within the \texttt{DreamFactory} framework, they function similarly 
to their real-world counterparts, taking on roles such as determining the movie's style, writing scripts, and drawing.
\begin{figure}
\begin{minipage}[t]{0.25\textwidth}
  \centering
  \includegraphics[width=1.0\linewidth]{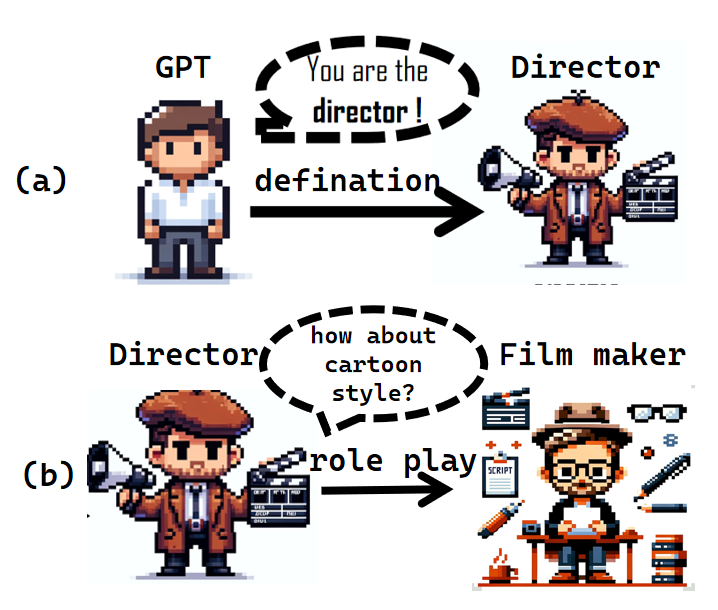} 
\end{minipage}%
\begin{minipage}[t]{0.25\textwidth}
  \centering
  \includegraphics[width=1.0\linewidth]{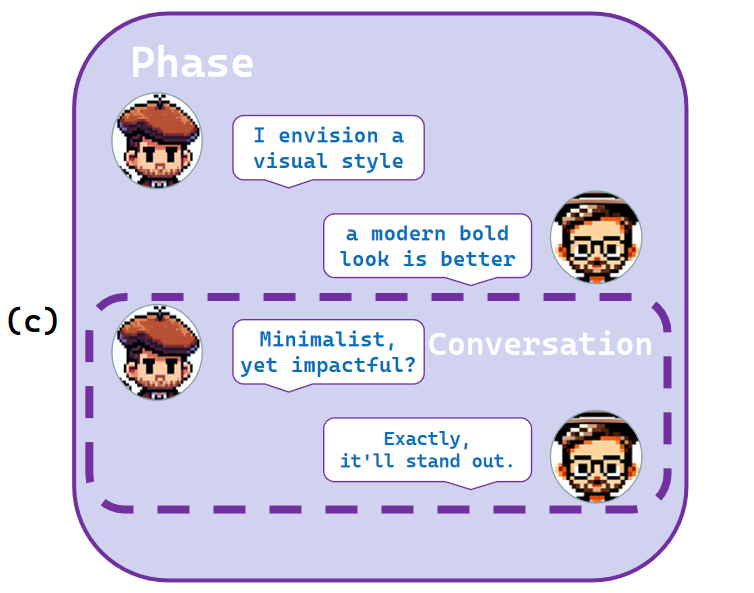}
\end{minipage}
\caption{The Figure demonstrates how GPT begins role-playing as a director and commences communication with other GPTs as a director would.}
\label{fig:GPT_roleplay}
\end{figure}

The definition prompts for their roles primarily consist of three main parts: 
\textbf{{Job}},
\textbf{Task} and 
\textbf{Requirements}. 
For instance, the definition prompt for a movie's creator would include the following sentences: (a) You are the \textbf{Movie Art Director}. Now, we are both working at Dream Factory,... (b) Your job is to \textbf{generate a picture according to the scenery} given by the director...and (c) you must \textbf{obey the real-world rules}, like color unchanged...
For tasks such as plot discussions, we also limit their discussions to not exceed a specific number of rounds (depending on the user's settings and the company's size definition). We have included the following prompt to ensure this: "You give me your thought and story, and we should brainstorm and critique each other's idea. After discussing more than 5 ideas, any of us must actively terminate the discussion by picking up the best style and replying with a single word <INFO>, followed by our latest style decision, e.g., cartoon style."

In Figure~\ref{fig:GPT_roleplay}, panels (a) and (b) feature schematic illustrations of a character being defined and 
initiating role play.
The complete architecture of the entire company is fully introduced in Figure~\ref{fig:roles_in_DF}.
For each, we defined a role card, which contains: 1) The role name is put on the left-upper corner
of each card; 2) The phases of the role involved are put on the right-upper corner of each card;
3) On each role card, we show the role-involved conversation and collaborative roles; 4) We show
the intermediate output of the role on the right-hand side of the card; and 5) Finally, we put the
corresponding files or content out of conversations on the bottom of the card.

\subsection{\texttt{DreamFactory} Framework pipeline}
In this section, we introduce the specific pipeline of \texttt{DreamFactory}. 
\textbf{Figure~\ref{fig:overview}} illustrates the main phases and indicates which agents engage in conversations. 
Before delving into our entire pipeline, it's essential to first outline its fundamental components: 
phases and conversations. As depicted in \textbf{Figure~\ref{fig:GPT_roleplay}} (c, a phase represents a complete stage that 
takes some textual or pictorial content as input. Agents, composed of GPT, engage in roleplay, discussion, and 
collaboration for processing, ultimately yielding some output. A conversation is a basic unit of a phase, with 
typically more than one round of conversation encompassed within a phase.
After a fixed number of conversations, a phase is approaching its conclusion, at which point \texttt{DreamFactory} will 
save certain interim conclusions generated within this phase that we wish to retain. For instance, in the Phase 
style decision, the final conclusion will be preserved. Furthermore, during subsequent phases, \texttt{DreamFactory} will 
provide the necessary precedents, such as invoking previous styles and scripts when designing keyframes later on.

Recently, large language models were found to have their capabilities limited by finite reasoning abilities, 
akin to how overly complex situations in real life can lead to carelessness and confusion. Therefore, the main idea of 
this framework, in the video domain, is to decompose the creation of long videos into specific stages, allowing 
specific large models to play designated roles and leverage their powerful capabilities in analyzing specific problems. 
Like a real-life film production company, \texttt{DreamFactory} adopts a classic workflow, starting with scriptwriting followed 
by drawing. Overall, the framework encompasses six primary stages: 
\textbf{Task Definition}, 
\textbf{Style Decision}, 
\textbf{Story Prompting}, 
\textbf{Script Design}, and 
\textbf{Key-frame Design}
. The specific method for the final stage, keyframe iterative design, will be 
introduced in the following section; it is used to maintain the consistency and continuity of images generated at 
various stages. In the first four phases, our roles are conversational.

In each phase, every agent shares a "phase prompt" that includes the following key points: our roles, our tasks, 
the conclusions we aim to draw, the form of our discussion, and some other requirements. Following this, each 
agent is further informed by its unique prompt about its role definition, as discussed in section 3.1. We 
can refer to the notation in Guohao Li's article[1] to define the collaboration process of agents within \texttt{DreamFactory}.
We refer to the assistant system prompt/message by Pa and that of the user by Pu. The system messages
are passed to the agents before the conversations start. Let F1 and F2 denote two large-scale autoregressive language 
models. When the system message is passed to those models respectively, we can get  $A \leftarrow F_{1}^{P_{A}}$ , \space $U \leftarrow F_{2}^{P_{U}}$
which are referred to as the assistant and user agents respectively. In continuation, we assume that the text 
provided by the user (instructor) at each instance is denoted as \textbf{It}, and the response given by the 
assistant is denoted as \textbf{At}. The Output at time step \textbf{t} alternating conversations between the two can be represented as: $\mathcal{O}_t = \left( \left( I_1, A_1 \right), \left( I_2, A_2 \right), \ldots, \left( I_t, A_t \right) \right)$.

\begin{figure}[ht]
  \centering
  \includegraphics[width=0.5\textwidth]{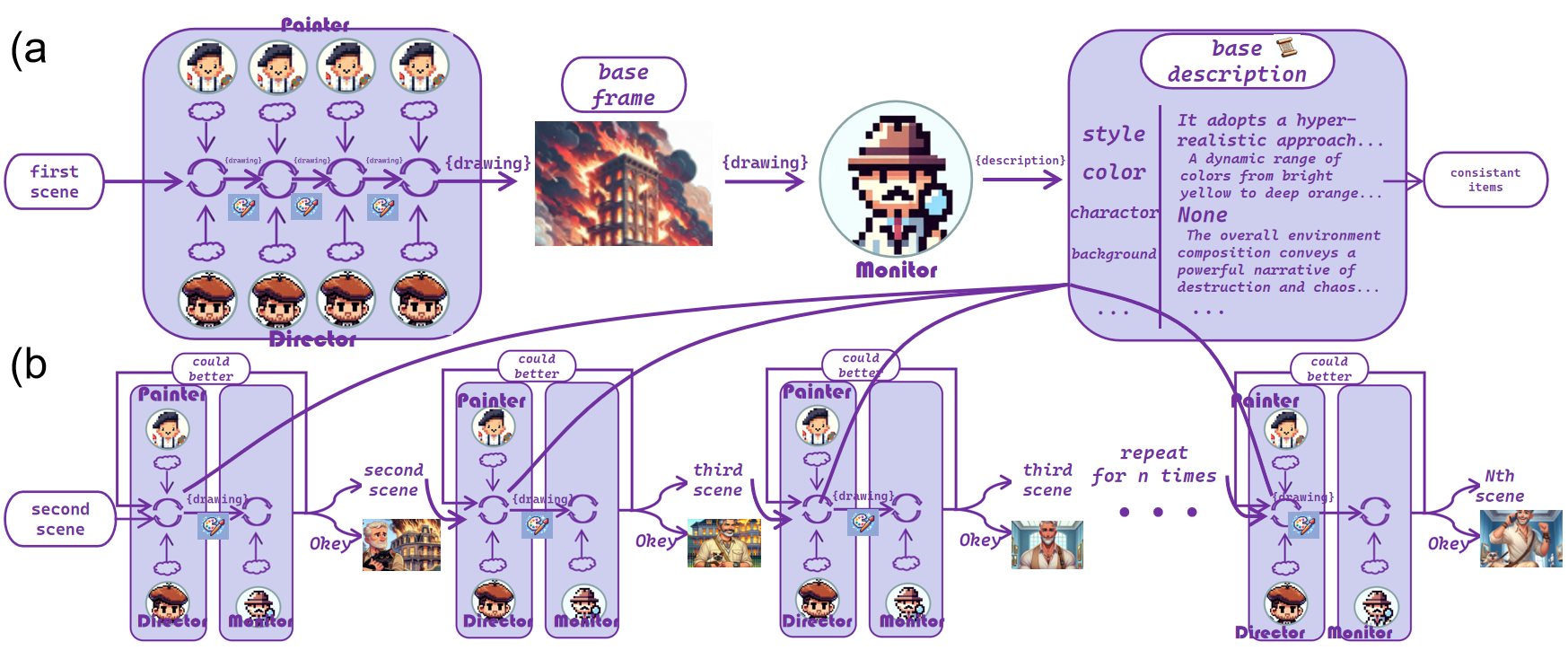}
  \caption{An overview of the keyframe iterative design. 
  }
  \label{keyframe}
\end{figure}

Following the five critical phases mentioned above, five significant outputs will be achieved. In the prompt, 
each phase's output \( O_t \) is required to follow <INFO> for summarization, which also allows us to systematically obtain and preserve, 
forming the Local memory information of the \texttt{DreamFactory} framework. This is also one of the primary purposes of proposing this 
framework, maintaining the consistency of critical information. Finally, after generating the tasks, styles, stories, scripts, and 
keyframe images, a long video with consistent style is obtained.

\subsection{Keyframe Iteration Design}

During the generation of long videos, the most challenging problem to address 
is that a video comprises a long sequence of image collections. Therefore, when 
generating, the model needs to maintain a long-term, consistent memory to ensure 
that each frame produced by the model coherently composes a consistent video. This 
type of memory includes two kinds: 
\textbf{short-term memory knowledge} and 
\textbf{long-term memory system}.

\textbf{short-term memory knowledge} is embedded within videos of a fixed scene. Between adjacent frames, the animation in each frame should be connected, the characters should be unified, and there should be no significant changes in color, style, etc. As of now, the latest video models perform very well in terms of short-term memory. Nonetheless, we have still added a Monitor to supervise whether our video model is performing sufficiently well. As illustrated in Figure~\ref{keyframe}, there is a review process after the generation of each frame. Therefore, to maintain short-term consistency, the supervisory mechanism we introduced has addressed this issue.

\textbf{long-term memory system}, however, pose a challenge that troubles most current models and represents the most pressing issue in video generation today. Particularly, within a GPT-based fully automated multi-agent framework, the inherent randomness and drift phenomena of large language models make this problem difficult to tackle. Long-term memory implies that across scene transitions, the model should be able to maintain the consistency of the drawing style, character continuity, and narrative flow. To uphold long-term memory, we have introduced the Keyframe Iteration Design method, which transforms long-term memory into short-term memory by guiding the generation of consecutive, consistent images, iterating and generating forward with each step.
\textbf{Figure~\ref{keyframe}} demonstrates the process of each iteration.

\textbf{Keyframe Iteration Design Method} leverages the inferential capabilities of large language models to transform 
long-term memory into iterations of short-term memory to ensure consistency. The first frame of the image is the beginning 
of the entire video and establishes essential information such as the style, painting technique, characters, and background 
for the entire long video. Therefore, we refer to the first frame as the Base. 
At the beginning, we will generate a painter $P$, a director $D$ and a monitor $M$, represented by $ P \leftarrow F_{1}^{P_{P}}$ ,  $D \leftarrow F_{2}^{P_{D}}$ ,  $M \leftarrow F_{3}^{P_{M}}$
,these models played by visual large language models, will engage in a cyclical process of generation and discussion 
until they produce a crucial frame, which is the first keyframe, referred to as the \textbf{Base Frame}. At this point, the Monitor $D$, composed 
of a visual large language model as well, will conduct a thorough analysis to extract information, detailed description of features such as 
style, background, and character traits that should be preserved for an extended period. This results in the \textbf{Base Description}, note as $B_{D}$. $S1$ represents the script for the first frame. We have $\mathcal{O}_t = Gen \left( 
\begin{array}{c}
 p_t, d_t ,S_1
\end{array}
\right)$, where $B_{D} \leftarrow M(O_{t})$.



In subsequent generations, when iterating the keyframe for moment $t$, we will use the previously input $S_{t}$ as the description of the 
scene. To maintain continuity in the context of adjacent scenes, we will employ the nurtured method to generate the description 
for the moment $t-1$, which we also refer to as the contextual environment denoted as $C_t-1$.
At the same time, to maintain long-distance memory, $B_{D}$ will also serve as an input. By referencing the basic features of the 
previous frame and the Base features, it can ensure that the necessary information is essentially grasped in the next iteration, 
enabling the drawing of continuous keyframes with the same style, consistent characters, and uniform background. We have $\mathcal{O}_t = Gen \left( 
\begin{array}{c}
p_t, d_t, S_t, C_{t-1} 
\end{array}
\right)$.


Upon the previous generation of keyframes, we can obtain the contextual environment and proceed with the next round of generation. We have $C_t = M(O_t)$, $  p_{t+1} = P(S_t , C_t)$ ,  $d_{t+1} = D(S_t , C_t , p_{t+1})$.
Ultimately, we achieve the generation of the keyframes for the moment $t+1$.

In practical application, controlling the details of characters proves to be the most challenging aspect. Therefore, under our carefully modified prompts, with increased emphasis on parts that performed poorly in multiple experiments, the Keyframe Iteration Method can now generate a very consistent and practically valuable series of images.

\section{Experiments}


\subsection{Traditional Video Quality Evaluati1on}
\textbf{Evaluati1on Metrics} - To validate the continuity of the keyframes 
and the quality of the videos produced by the framework, we embedded various 
tool models (such as Runway, Diffusion, GPT) within the architecture to 
assess the quality of videos generated by different tools. In our 
experiments, we principally employed the following evaluation metrics: 
\textbf{(1)} Fréchet Inception Distance (FID) score: measures the similarity between generated images and real images.
\textbf{(2)} Inception Score (IS): gauges the quality and diversity of generated images.
\textbf{(3)} CLIP Score: evaluates the textual description accuracy of generated images.
\textbf{(4)} Fréchet Video Distance (FVD) score: extension of the FID for videos, comparing the features distribution of real videos versus synthesized ones based on Fréchet distance
and 
\textbf{(5)} Kernel Video Distance (KVD): utilizes kernel function to compare the features distribution of real videos versus synthesized ones.

Our dataset, during the Regular phase, comprised conventional prompts consisting of 70 keywords and brief sentences randomly selected by experimental personnel from the COCO dataset. This was utilized to evaluate the generated image quality of the fundamental tool models and the degree of alignment between the images and the text. For the Script phase, scripts pertaining to 70 randomly extracted tasks from our provided dataset were employed during the script-filling stage. This guided the model generation based on the relevant plot to assess the function of the "Animation Department" within the \texttt{DreamFactory} framework. The \texttt{DreamFactory} label denotes the keyframe images produced by the framework that corresponds to the Script.\looseness=-1









\textbf{Output Quality Statistics} - The images generated using models such as DALL·E and Diffusion are of high quality and have reached the state-of-the-art level in various indices. To quantitatively analyze the quality of the generated images, we input the images corresponding to the original prompts into GPT  to get the GPT-Script and then used original prompts or the GPT-Script as prompts to generate 1400 images, from which we calculated FID, IS, and CLIP Score. As for FVD and KVD, we selected 100 samples from our multi-scene video dataset and manually extracted 10 keyframes for each one, Which can be used to generate multi-scale videos.

Data in Table~\ref{table1} indicates that the quality of images generated using scripts is on average more refined than those produced using everyday prompt words. This may be attributable to the extent to which GPT acts as a prompt, and contemporary models are generally adept at processing longer prompts. However, within the \texttt{DreamFactory} framework, the application of keyframe iterative design, in conjunction with storyboard creation, detailed descriptions of characters, settings, lighting, and style determination, has led to a marked improvement in the quality of image generation. A similar enhancement is also evident in videos which is shown in Table~\ref{tab:my_label}.
\begin{table}[ht]
  \centering
  \resizebox{\linewidth}{!}{
  \begin{tabular}{lccc} 
  \hline

  \hline

  \hline

  \hline
  \textbf{Models Composition} & \textbf{FID} & \textbf{IS}  & \textbf{CLIP Score}   \\ \hline
  Dalle-e3 (Regular) &  9.30 &  133.46 & 26.69\\
  Diffusion (Regular) & 9.15 & 158.23 &  26.58  \\
  Midjourney (Regular) & 11.23 & 163.20 & 25.91   \\
  GPT3.5-Script+Dalle-e3 & 9.78 & 153.43  & 29.58  \\
  GPT3.5-Script+Diffusion &  8.63 & 168.90 & 30.57 \\
  GPT3.5-Script+Midjourney &  10.81  & 174.45  & 29.32 \\
  GPT4-Script+Dalle-e3  &  8.53 &  159.12  &   29.84 \\
  GPT4-Script+Diffusion  & 8.32  &  169.97  & 30.73  \\
  GPT4-Script+Midjourney &  10.26  & 178.14    &  29.75\\ 
  \texttt{DreamFactory}(GPT4)+Dalle-e3  & \textbf {6.57}  &  \textbf {160.94}  & \textbf{30.76}  \\
  \texttt{DreamFactory}(GPT4)+Diffusion  & \textbf{7.03} &  169.71  & \textbf{30.92}\\ 
  \texttt{DreamFactory}(GPT4)+Midjourney  &\textbf{7.15}& \textbf{178}  &    
  \textbf{30.39} \\ \hline

  \hline

  \hline

  \hline
  \end{tabular}}
  \vspace{0.2 cm}
  \caption{The statistical analysis of Text2Image task. All models can generate higher-quality images after prompts augmentation, but the quality of the images generated by our framework stands out. }
  \label{table1}
\end{table}

\begin{table}[ht]
    \centering
    \resizebox{0.8\linewidth}{!}{
    \begin{tabular}{lcc}
    \hline

  \hline

  \hline

  \hline
    \textbf{Models Composition} & \textbf{FVD} & \textbf{KVD} \\ \hline
    Runway (Regular)  &  1879  &  125 \\
    Stable Video (Regular) & 3560  & 182 \\ 
    \texttt{DreamFactory}+Runway  &  \textbf{732}  & \textbf{62}   \\
    \texttt{DreamFactory}+Stable Video  & \textbf{1376}  &  \textbf{113} \\\hline

  \hline

  \hline

  \hline
    \end{tabular}}
    \vspace{0.2 cm}
    \caption{The statistical analysis of Image2Video task. The improvement of our framework for generating multi-scene long videos is remarkable. }
    \label{tab:my_label}
\end{table}

\subsection{Multi-scene Videos Evaluation Scores}
\textbf{Cross-Scene Face Distance Score} - In the generation of sequential videos, addressing character consistency is paramount. 
Discrepancies in the appearance of characters can lead not only to poor visual perception but also to the audience's 
inability to understand the plot and content. Maintaining character consistency ensures the coherence of the storyline 
revolving around the characters and enhances the visual appeal of the video.
Especially, in the domain of long-duration videos, a video is typically composed of multiple scenes. 
This represents an unprecedented area of research, where there is a pressing need for robust evaluation metrics 
to assess the consistency of characters appearing across complex, multi-scene videos. Against this backdrop, 
we experimentally introduce the concept of the Cross-Scene Face Distance Score\textbf{(CSFD Score)}, aimed at validating the issue of
character facial feature consistency across different scenes.

\begin{figure}[h]
  \begin{minipage}[c]{0.47\textwidth}
    \includegraphics[width=\textwidth]{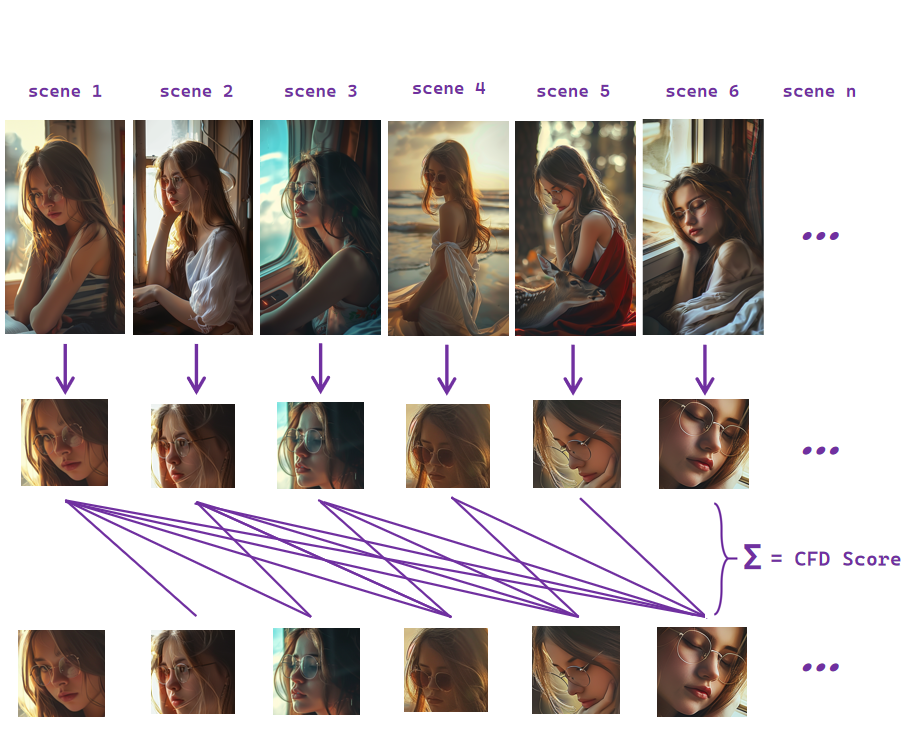}
    \caption{Schematic diagram and pseudocode for the calculation of Cross-Scene Face Distance 
Score. 
}
\label{pse1}
  \end{minipage}\hfill
  \begin{minipage}[c]{0.5\textwidth}
  \begin{algorithm}[H]
  \captionof{algorithm}{Calculate CSFD Score}
          \label{a1}
    \begin{algorithmic}[1]

        \State $total \gets 0$
        \State $count \gets \text{n*(n-1) / 2}$
        \For{$i \gets 1 \text{ to } n$}
            \For{$j \gets i+1 \text{ to } n$}
                \State $similarity \gets \text{CFS}(F_i, F_j)$
                \State $total \gets total + similarity$
            \EndFor
        \EndFor
        \State $averageScore \gets total / count$
        \State \Return $averageScore$
    \end{algorithmic}
    \end{algorithm}

  \end{minipage}
\end{figure}

In the computational process, each keyframe corresponds to a face, and using the dlib library, the position of the face can be extracted. The face-recognition library
can be used to calculate the similarity score. For the facial segment of each frame, 
we can compute its similarity with all subsequent frames and then take the average. By this method, 
we can accurately determine whether the faces in the video are consistent.
The relevant schematic diagram and the pseudocode for the calculation are provided in Algorithm~\ref{a1}.

\textbf{Cross-Scene Style Consistency Score} - In the production of long videos, maintaining stylistic consistency is equally important. A consistent 
style makes the video appear as a cohesive whole. Based on this concept, we have introduced the Cross-Scene 
Style Consistency Score\textbf{(CSSC Score)}. However, to my knowledge, there currently isn't a mature method to rapidly determine the style 
of a video, so at this stage, we will rely on the assistance of large language-visual models. Essentially, 
we divide the video into several categories, which include:\textbf{ anime, illustration, origami, oil painting, 
realism, cyberpunk, and ink wash}.

\begin{figure}[h]
  \begin{minipage}[c]{0.5\textwidth}
      \includegraphics[width=\textwidth]{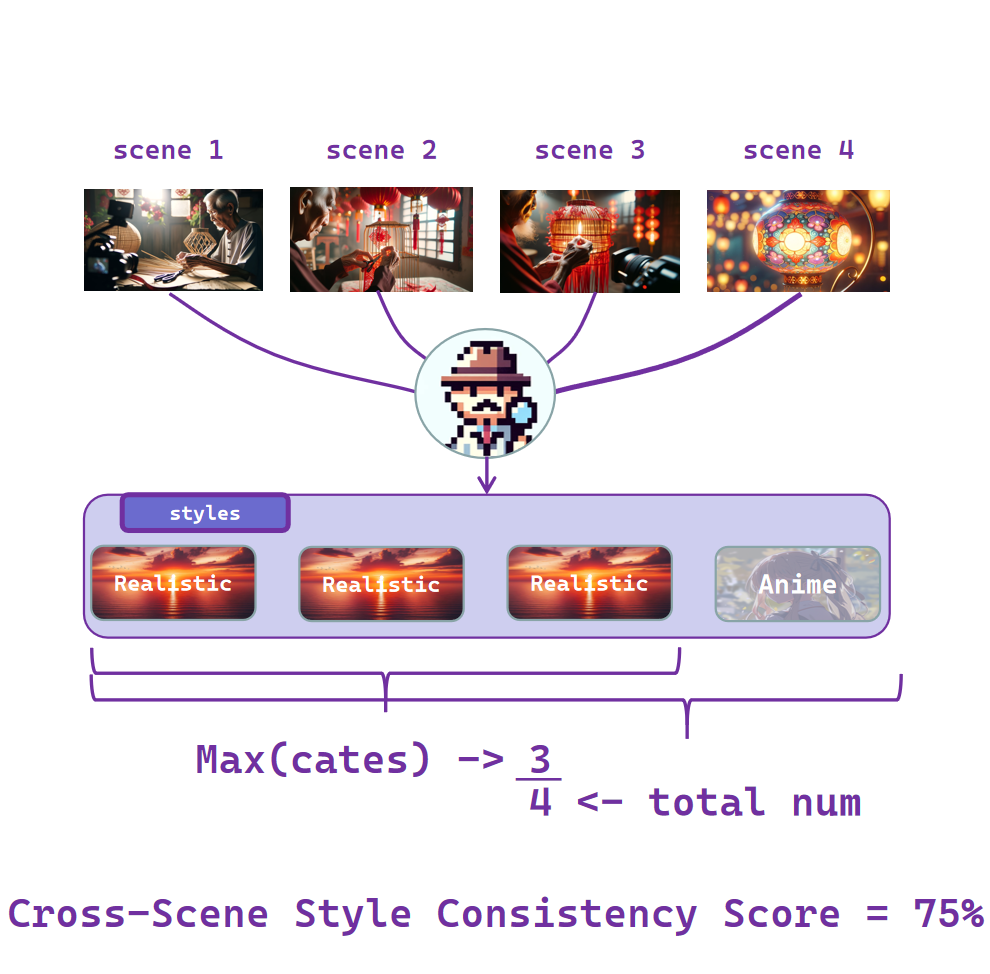}
      \caption{Schematic diagram and pseudocode for the calculation of Cross-Scene Style Consistency Score. }
      \label{pse2}
  \end{minipage}\hfill
  \begin{minipage}[c]{0.5\textwidth}
    \begin{algorithm}[H]

      \caption{Calculate CSSC Score}
          \label{a2}
      \begin{algorithmic}[1]
      \State $n \gets \text{number of key frames}$
      \State $categories \gets \text{array initialized to 0 of size number of categories}$
      \For{$i \gets 1 \text{ to } n$}
          \State $category \gets \Call{Judge}{F_i}$
          \State $categories[category] \gets categories[category] + 1$
      \EndFor
      \State $maxCount \gets \max(categories)$
      \State $crossSceneStyleScore \gets \frac{maxCount}{n} \times 100$
      \State \Return $crossSceneStyleScore$
      \end{algorithmic}
      \end{algorithm}

  \end{minipage}
\end{figure}

The calculation method for the Cross-Scene Style Consistency Score is as follows: For each key frame, a divider 
played by a GPT-4V is used to determine the classification. Once all scenes have been clearly divided 
into categories, the proportion of the most numerous category to the total number of key frames is 
calculated. Figure~\ref{pse2} presents a partial output where the input is "an elderly person making a traditional 
Chinese lantern in real life". Scene 4 depicts an animated lantern created using Dalle, with GPT-4V serving 
as the discriminator. It is observable that among the four scenes, the first three are categorized under a 
realistic style, while the fourth scene is classified as anime style. Consequently, the maximum number of 
distinct styles is three, resulting in a cross-scene style consistency score of 75\%.
The other relevant schematic diagram and the pseudocode for the calculation are provided in Algorithm~\ref{a2}.

\begin{table}
  \centering  
\resizebox{\linewidth}{!}{
  \begin{tabular}{lccc} 
  \hline

  \hline

  \hline

  \hline
  \textbf{Models} & \textbf{CSFD Score}&  \textbf{CSSC Score} & \textbf{av-Clip Score}\\ \hline
  GPT4-Script+Dalle-e3 & 0.77& 0.85& 0.29\\
  GPT4-Script+Diffusion& 0.75& 0.83& 0.28\\
  GPT4-Script+Midjourney& 0.68& 0.66& 0.26\\
  \texttt{DreamFactory}(GPT4)+Dalle-e3 & \textbf{0.89}  & \textbf{0.97} & \textbf{0.31} \\ \hline

  \hline

  \hline

  \hline
  \end{tabular}}
  \vspace{0.2 cm}
  \caption{The statistical analysis of cross-scene score on different models.}
  \label{table3}
\end{table}
\textbf{Average Key-Frames CLIP Score} - In the generation of long videos with multiple scenes, it is crucial to assess the alignment of each scene's keyframes with the 
corresponding text. They have incorporated a significant amount of additional information to ensure consistency, which could likely 
lead to deviations from the text during generation. This may result in the overall video not adhering to the script. Therefore, in this 
section, we propose the Average Key-Frames CLIP Score to ensure the consistency of key frame scenes with the script. The calculation 
method is straightforward: compute the CLIP score for each keyframe against the scene generated during scene prompting and take the 
average.

\textbf{Results} - In table~\ref{table3}, our data selection comprised seventy character-centric entries from the Multi-Scene Videos Dataset, produced by \texttt{DreamFactory} + GPT-4 + DALL-E 3. The baseline utilizes the DALL-E 3 model with script inputs from this segment. Furthermore, evaluations were conducted on the aforementioned (1) cross-scene facial distance, (2) cross-scene style scores, and (3) average CLIP Score. These metrics were used to assess the consistency of facial features within our framework, the consistency of scene attributes, and the alignment between prompts generated by our framework and the narrative, as well as imagery.

\begin{figure}[h]
  \centering
  \includegraphics[width=0.5\textwidth]{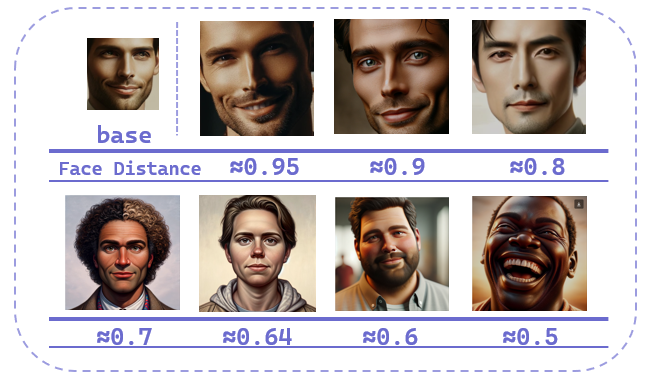}
  \caption{The distance between different faces when using openai-clip as the encoder.} 
  \label{dis}
\end{figure}

In our Cross-Scene facial distance scoring experiment, we employed the face locations 
method from the face-recognition library to locate 68 facial landmarks, thereby 
focusing the portrait photographs on the facial area. During the image encoding phase, 
we utilized the ViT model from the openai-clip repository to input the facial region 
and compute the vector representations. Subsequently, a vector dot product operation 
was performed to determine the final facial distance score. Owing to the inherent 
similarity among the facial images, all the scores were predominantly above 0.5. The 
specific reference facial match-score pairs are exhibited in Figure~\ref{dis}. In the analysis 
of both the CSSC score and the average CLIP score, the same set of seventy random 
samples was utilized as data. The CSSC Score employed GPT-4 Version as the stylistic 
analyst.









\section{Conclusion}
We introduce \textbf{Dream Factory}: a multi-agent-based framework for generating long videos with multiple scenes. 
Dream Factory incorporates the idea of multi-agents into the field of video generation, producing consistent, continuous, and 
engaging long videos. Dream Factory introduces a 
keyframe iteration method to ensure alignment of style, characters, and scenes across different frames and can be built on 
top of any image or video generation tool. Furthermore, Dream Factory proposes new metrics to validate its capabilities by 
measuring the quality of generated content through cross-scene face and style consistency, as well as text-to-visual alignment. On the test set, the DreamFactory framework can achieve highly consistent sequential story generation, marking a groundbreaking development.


\clearpage
\section{Limitations}
In this paper, we present a multi-agent video generation framework capable of producing videos with high consistency across multiple scenes and plotlines. However, we still face several limitations. Firstly, our current reliance on prompts to control agents means that the agents are not capable of highly creative tasks, such as devising plots with artistic merit. Such tasks require the accumulation of specific datasets for model fine-tuning. Secondly, the editing of all video segments is centered around synthesized speech content, which results in a final product that may appear as a mere assembly of clips. This necessitates the introduction of a unique framework design to enhance the fluidity of the videos. Lastly, video generation still involves substantial resource consumption.
\section{Ethics Statements}
The development and deployment of DreamFactory, a multi-agent framework for long video generation, raise several ethical considerations that must be addressed. The potential for the misuse of generated videos, such as the creation of deepfakes or the propagation of misinformation, is a significant concern. To mitigate these risks, we commit to implementing robust safeguards, including watermarking generated content and collaborating with fact-checking organizations. Additionally, we will ensure transparency in our research and make our methods and datasets publicly available, subject to ethical use guidelines. We also recognize the importance of diversity and inclusion in the training data to prevent biases in the generated content. Finally, we will engage with the broader community to establish ethical standards for the use of AI-generated video content, promoting responsible innovation and use of this technology.

\bibliography{ref}

\clearpage
\appendix

\section{Appendix}
\label{sec:appendix}
\subsection{DreamFactory Responsibility allocation}

\begin{figure*}[ht]
  \centering
  \includegraphics[width=1.0\textwidth]{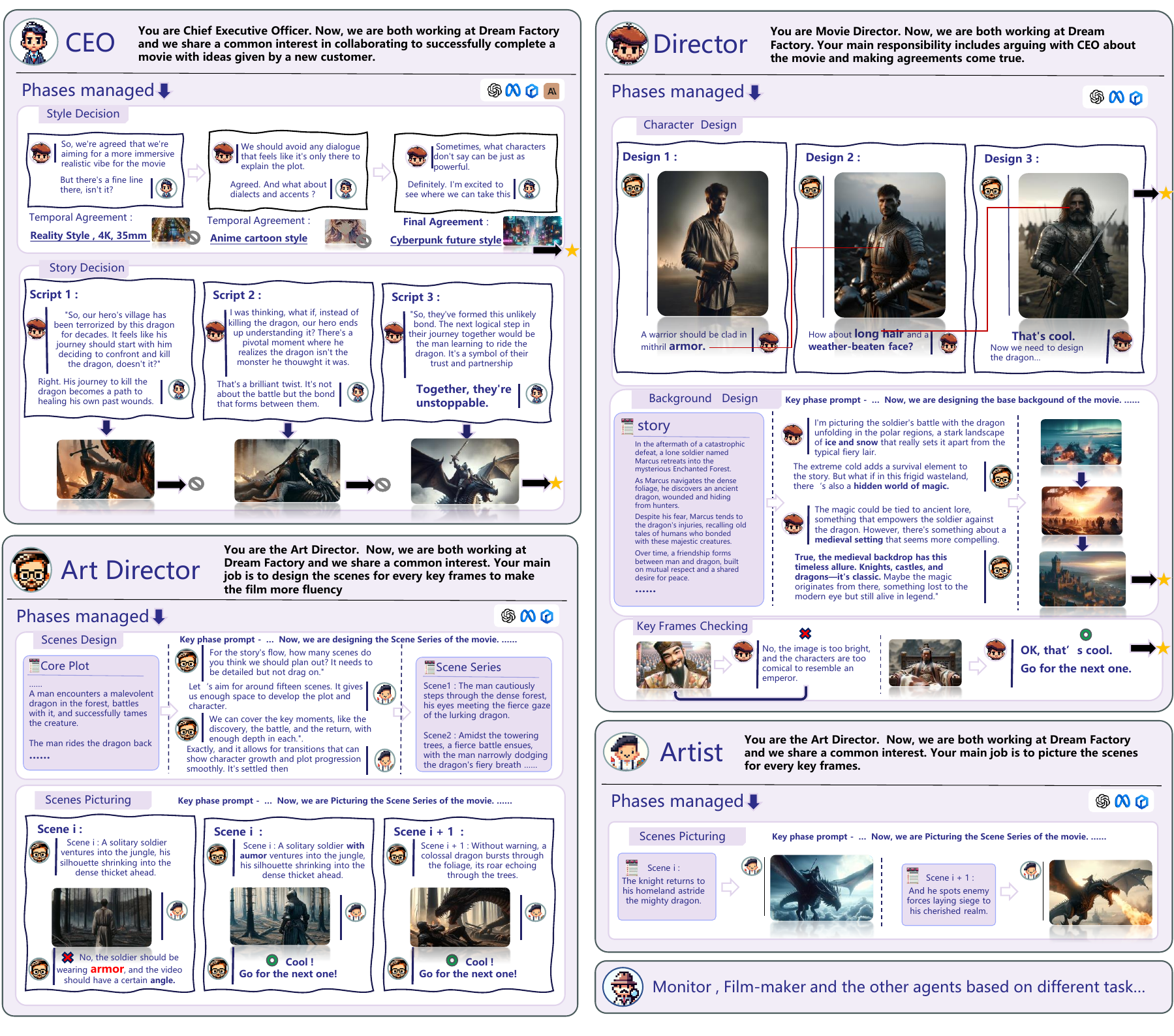}
  \caption{This figure presents the responsibility allocation chart for all employees within the \texttt{DreamFactory} architecture. For each employee, the upper left corner displays their role and portrait, while the upper right corner outlines the stages of participation and their roles. The essential parts of the prompt are depicted below.}
  \label{fig:roles_in_DF}
\end{figure*}

As shown in Figure~\ref{fig:roles_in_DF}, our 
\texttt{DreamFactory} framework utilizes multiple large language models (LLMs) to form a simulated animation company, 
taking on roles such as CEO, Director, and Creator. Given a story, they collaborate and create a video through social 
interaction and cooperation. This framework allows LLMs to simulate the real world by using small video generation 
models as tools to accomplish a massive task. As illustrated in Figure~\ref{fig:roles_in_DF}, under their collaboration, it is possible to 
generate a series of consistent, stable, multi-scene long videos as the plot progresses.

\subsection{User Study}
Quantitative evaluation of human preference for video is a complex and difficult proposition, so we employed human evaluators to verify the quality of multi-scene videos generated by our framework. We collected 150 multi-scene short videos generated by AI from the internet and compare them with videos from our framework. Through this approach, we aimed to assess whether our videos could achieve an advantage in human preferences compared to existing AI videos on the network.

\begin{figure}
  \centering
  \includegraphics[width=0.5\textwidth]{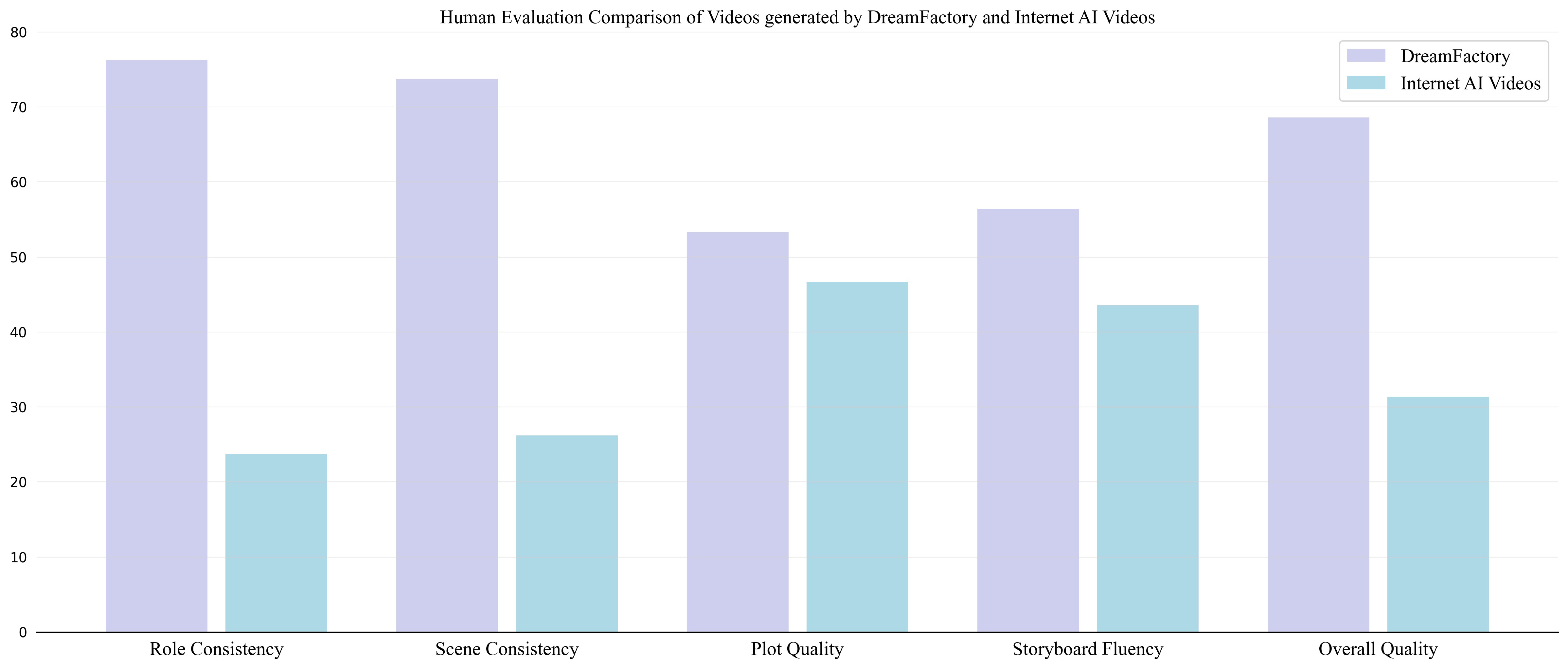}
  \caption{Human evaluation comparison of videos generated by \texttt{DreamFactory} and internet AI videos.}
  \label{ima}
\end{figure}

In our study, We adopt the Two-alternative Forced Choice (2AFC) protocol, as used in previous works [\cite{blattmann2023stable},~\cite{blattmann2023align}, ~\cite{bartal2024lumiere}].  In this protocol, each participant will be randomly shown a pair of videos with the same story, one is a short video collected on web platforms and the other is generated by our framework. Participants were then asked to select the superior side on five metrics: role consistency, scene consistency, plot quality, storyboard fluency, and overall quality.  We collected 1320 human scores for this study, utilizing schools, communities, and network platforms. As illustrated in Figure~\ref{ima}, our method was preferred better.

\subsection{Case Study}
\begin{figure}
  \centering
  \includegraphics[width=0.5\textwidth]{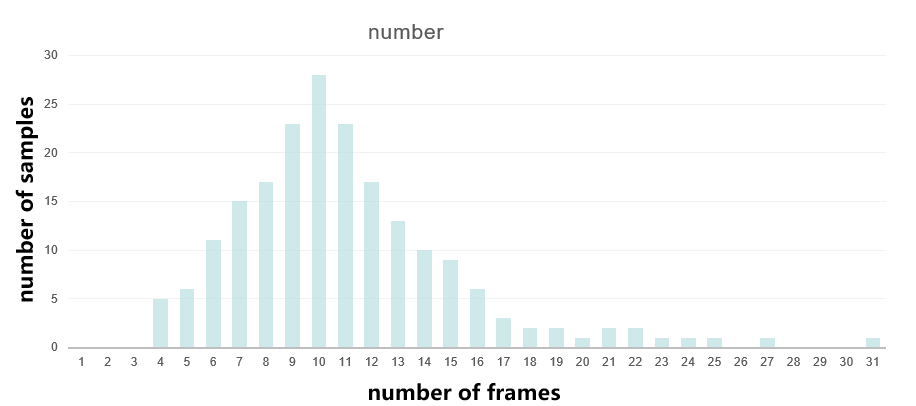}
  \caption{The key frame numbers count Statistics of \texttt{DreamFactory}.}
  \label{fig:framesnumberstatic}
\end{figure}

\textbf{Comprehensive Keyframe Count Statistics} - The version currently provided to users is balanced between cost and user experience, using the Short generation mode, typically around ten scenes. The specific number is related to the user's task input. The length of videos generated using random prompts is shown in the figure \ref{fig:framesnumberstatic}.



\end{document}